Bessou Sadik, Louail Mohamed, Refoufi Allaoua
Kadem Zehour& Touahria Mohamed
Faculté des sciences de l'ingénieur, Département d'informatique, Université de Ferhat Abbess, Algérie

**Un système de lemmatisation pour les applications de TALN**

Cet article propose une méthode de lemmatisation (stemming) pour les textes arabes, basée sur les techniques linguistiques de traitement du langage naturel. Cette méthode s'appuie sur la notion de schème (une des points forts de la morphologie de la langue arabe). L'avantage de cette approche est qu'elle n'utilise pas de dictionnaire des formes fléchies mais qu'elle effectue une reconnaissance dynamique et intelligente des différents mots de la langue.

**Mots clés**
Lemmatisation, analyse morphologique, TALN, langue arabe

**Abstract**
This paper presents a method of stemming for the Arabian texts based on the linguistic techniques of the natural language processing. This method leans on the notion of scheme (one of the strong points of the morphology of the Arabian language). The advantage of this approach is that it doesn't use a dictionary of inflexions but a smart dynamic recognition of the different words of the language.

.
**I. Introduction**
    La masse d'information véhiculée à travers les réseaux est de plus en plus profuse et incontrôlable. La tendance est celle d'une surinformation tentaculaire, et le problème de la maîtrise de l'information risque de devenir très délicat voire nocif à moins que des outils de collecte, de traitement, de diffusion ciblée et d'exploitation de l'information ne viennent aider les utilisateurs à mieux gérer ces avalanches d'informations**[16]**. Dans ce cadre,  cet article expose un procédé de lemmatisation dans l'analyse morpho-lexicale. Cette étape est cruciale car elle est la base de tous les applications de TALN (Traitement Automatique des Langues Naturelles): traduction automatique, indexation, recherche d'information, réponses aux questions, …)  et aussi de la fouille de texte ou du *text mining* "finding interesting regularities in large textual datasets…", base pour la classification et la catégorisation des documents.
 L'application Du traitement automatique sur la langue arabe pose des problèmes majeurs **[6] [8]**, dont : le problème de l'ambiguïté issue de l'absence des voyelles **[1]**, ceci exige des règles morphologiques complexes **[11]**

    Le problème de reconnaissance des formes fléchies, car l'arabe est une langue fortement flexionnelle.**[5]**

**Lemmatisation**

C'est une procédure ramenant un mot portant des marques de flexion (par exemple, la forme conjuguée d'un verbe) à sa forme de référence (dite *lemme*), quelle que soit la forme sous laquelle le mot apparaît dans un texte. La lemmatisation sert ainsi à la reconnaissance morphologique des mots d'un texte.

*"Stemming forms of the same word are usually problematic for text data analysis, because they have different spelling and similar meaning so stemming is a process of transforming a word into its stem (normalized form)".* **[12]**

## 2. Analyse morphologique

L'analyse morphologique a pour but de vérifier si un mot fait partie de la langue traitée ou non. Elle consiste à décomposer les mots en morphèmes **[4]**, **[17]** sans tenir compte des liens grammaticaux entre ces derniers.

On peut souligner que l'analyse envisagée ici diffère de l'analyse lexicale classique en compilation, car elle ne se basera pas sur un automate, mais sur certaines notions concernant la morphologie des entités de la langue à analyser.

### 2.1. Principe de l'analyseur

Il procède en premier lieu par une normalisation qui transforme le document dans un format plus facilement manipulable **[14]**, elle est nécessaire à cause des variations qui peuvent exister lors de l'écriture d'un même mot puis la segmentation découpe le mot pour que chacune des parties obtenues soit une entité lexicale.
Cette segmentation isolera les préfixes et les suffixes du mot. La partie restante correspondra à la *racine* dans le cas où la segmentation est poussée jusqu'à la fin (par l'utilisation de la notion du schème). Dans le cas contraire la partie restante est appelée une *base*. Cette étape est une tache délicate du fait que l'arabe est une langue flexionnelle et fortement dérivable **[3]**

L'analyseur morphologique ne peut fonctionner sans l'aide d'un dictionnaire contenant les unités lexicales. Cette étape est l'analyse lexicale qui permet de vérifier si l'unité lexicale appartient bien à la langue, mais qui doit aussi vérifier la compatibilité entre les différents constituants du mot. Une troisième étape interprète les différents éléments obtenus par la segmentation en se basant toujours sur la notion de schème, et retourne comme résultat les valeurs morphologiques (hors contexte) des différentes unités lexicales, et probablement intervention des règles micro syntaxiques. Le schéma qui suit récapitule les étapes de l'analyse.

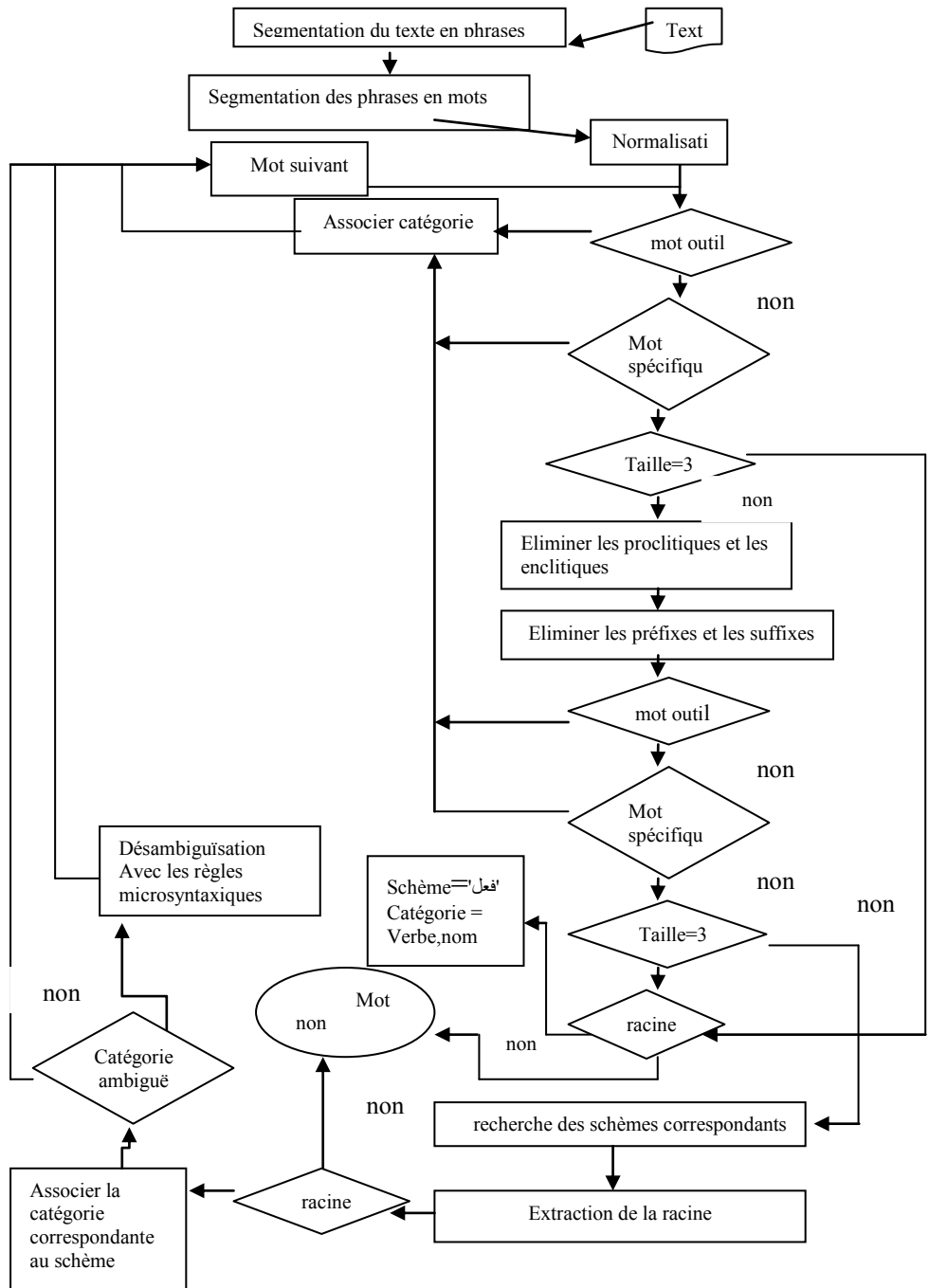

**Figure 1:** Schéma général de l'analyse morphologique

*2.2. Les dictionnaires nécessaires*
*2.2.1. Dictionnaire des schèmes*

Dans ce dictionnaire, on trouve toutes les informations qui seront utiles lors de la codification des mots sous forme de structure de traits. Il faut spécifier pour chaque schème son type: schème de nom ou de verbe.

On trouve aussi dans ce dictionnaire une codification des schèmes pour faciliter la recherche du schème et de la racine de l'entité comme il va être expliqué ci-après.

*2.2.2. Dictionnaire des racines*

Dans ce dictionnaire sont stockées toutes les racines représentatives de la langue, dans notre système les racines les plus utilisées dans la langue arabe. Ceci nous permettra de vérifier la compatibilité entre la racine et le schème, évitant ainsi toute décomposition erronée.

*2.2.3. Dictionnaire des mots-outils*

Nous considérons comme mot-outil, tout mot qui reste invariant quel que soit son contexte (à l'exception des noms propres et mots communs) tel que les pronoms, particules,…etc. Nous ne mettons dans ce dictionnaire que les mots-outils isolés.

Ce dictionnaire contient les prépositions, les particules qui ont un effet de rection sur les verbes à l'accompli et à l'inaccompli, les particules de coordination et d'autres particules. Il faut préciser que les particules vont être divisées dans quelques mots outils, des spécifications qui aident dans l'analyse syntaxique **[18].**

**Remarque**

Les spécifications mises dans les dictionnaires : schème, racine, mot-outil, sont les traits qui seront utilisés dans la partie analyse syntaxique.

*2.2.4. Dictionnaire des mots spécifiques*

Les mots de la langue arabe ne sont pas tous décomposables en racine et schème. Il existe en plus des mots-outils, une catégorie dite de mots spécifiques qui répondent à ce type, et qui est formée de mots qui n'ont pas une origine arabe, de noms propres et communs. La création d'un dictionnaire de ces mots spécifiques s'avère nécessaire.

Pour les noms communs et noms propres, un fichier est créé en laissant à l'utilisateur la possibilité de rajouter les noms qui manquent d'une part, d'autre part il suffit de considérer les décompositions mises en échec comme étant nom propre ou commun.

*2.2.5. Dictionnaire des mots vides du domaine*

Ce dictionnaire est réservé pour les mots les moins porteurs de sens dans un domaine donné (non pertinents)

*2.3. Structure des dictionnaires*
*2.3.1. Structure du dictionnaire des schèmes*

Rappelons que le schème utilisé par les grammairiens arabes comme point de départ de toute morphologie de la langue est 'فعل'. Par conséquent, à chaque mot du lexique arabe est associé un schème qui est le même mot sauf les lettres de sa racine qui sont remplacées par les lettres de la racine 'فعل'.

**Exemple**

mot = 'صالح' schème = 'فاعل'.

Le dictionnaire des schèmes est structuré de la façon suivante :

| Wzn i | Listeinfixe i | Catégorie |
|-------|---------------|-----------|
| افتعل | 13            | Verbe     |

**Figure 2** : structure du dictionnaire des schèmes.

Le champ *Wzn* contient la chaîne consonantique du schème. Le champ listeInfixe i contient la position des lettres autres que les lettres de la racine dans le schème et le champ catégorie donne la catégorie grammaticale (nom, verbe, .....).

Cette structure est adaptée pour faciliter la recherche des schèmes qu'on va présenter dans la suite.

*2.3.2. Structure du dictionnaire des racines*

| Asl |
|-----|
| .......... |
| خرج |
| …….. |
| هرب |

**Figure 3:** structure du dictionnaire des racines.

Le champ Asl contient la chaîne consonantique de racine. Les racines sont insérées par ordre alphabétique. Il est composé toujours de trois caractères.

**Remarque**

Il existe dans la langue arabe des verbes quadrétaires comme دحرج، زلزل
Mais, Notre système ne traite pas c
es exceptions car ils sont rares.

*2.3.3. Structure du dictionnaire des mots outils*

| Hrf | Classe |
|---|---|
| في | Nominale |
| قد | Verbale |
| و | Commune |

**Figure 4:** structure du dictionnaire des mots outils.

Le champ Hrf contient la chaîne consonantique du mot outil, le champ classe détermine le type du successeur du mot outil en précisant s'il est un verbe ou nom ou commun entre les deux.

*2.3.4. Structure du dictionnaire des mots spécifiques*

| Jmd |
|---|
| أحمد |
| سطيف |
| كمبيوتر |

**Figure 5:** structure du dictionnaire des mots spécifiques.

Le seul champ Jmd contient la chaîne consonantique du mot spécifique.

*2.4. Méthodologie utilisée*

Les grammairiens arabes s'accordent à dire que tout lexème arabe à l'exception des noms propres et de quelques noms communs, et des mots outils est le résultat de la combinaison d'une racine et d'un schème spécifique. On peut schématiser, l'extraction du schème et racine à partir du mot: مساجد (mosquées) par le schéma suivant :

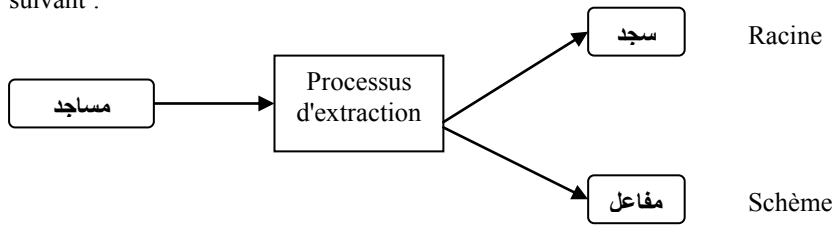

**Figure 6:** extraction du schème et racine.

Pour vérifier qu'un mot appartient aux lexèmes arabes il suffit de lui trouver sa racine et le schème correspondant suivant cette méthode. Le travail proposé va se baser sur trois étapes essentielles:

- Le découpage(segmentation);
- La recherche des schèmes et des racines;
- L'interprétation.

*2.4.1. Découpage*
Découper un mot ou le segmenter consiste à extraire ses différentes parties (préfixe, racine, suffixe,…).

*2.4.1.1. Techniques de segmentation*
　Résumons quatre techniques parmi d'autres.
**a) Première technique [19]**
　Elle consiste à découper le mot en trois éléments :
$$\text{proclynique + base + enclynique.}$$

**b) Deuxième technique [20]**
　Elle consiste à découper le mot en 5 éléments
$$\text{proclitique + préfixe + base + suffixe + enclitique.}$$

**c) Troisième technique [15]**
　Elle découpe le mot en : préfixe + racine + infixe + suffixe. Une procédure vérifie l'existence de la chaîne "ال" dans le mot. Si la chaîne existe alors le mot sera considéré comme un nom ou un adjectif, et la décomposition se fera avec les règles morphologiques des noms ou des adjectifs, sinon avec la décomposition des règles des verbes et des noms et des adjectifs.

**d) Quatrième technique [13]**
　Elle décompose le mot en Préfixe + base + suffixe, la liste des préfixes et suffixes est la combinaison de toutes les lettres additionnelles (greffes).

*2.4.1.2. Comparaison des différentes techniques*
**Table 1**: comparaison des techniques de segmentation.

| Méthodes | Avantages | Inconvénients |
|---|---|---|
| **[13]** | Simplicité. | Trop d'ambiguïté. Beaucoup de tests inutiles. |
| **[15]** | Diminuer le nombre de tests dans le cas où il trouve la chaîne"ال". | Ambiguïté si la chaîne "ال" appartient aux radicaux. Si la chaîne "ال" n'appartient pas au mot alors perte de temps. La suppression des voyelles longues risque la perte des radicaux. |
| **[20]** | Moins d'ambiguïté. | La nécessité de plusieurs tables de compatibilité. Ce qui complique le module de découpage. |
| **[19]** | Nombre de constituants limités à 3 ce qui accélère le découpage. | Difficile de former tous les enclyniques et proclyniques de la langue arabe c'est-à-dire trop d'ambiguïté. |

Ce tableau comparatif nous amène à choisir la troisième technique **[20]** car elle extrait des racines avec leurs valeurs morphologiques avec moins d'ambiguïté. Pour résoudre l'ambiguïté, Aljlayl et Frieder montrent que la lemmatisation légère (approche basé sur suppression de suffixe et de préfixe) surpasse significativement celle basée sur détection de racine dans le domaine de recherche d'information **[2]**.pour notre cas nous avons considérer la lemmatisation légère qui consiste à déceler si des préfixes ou des suffixes ont été ajouté au mot **[9]**

Le principe du découpage est le suivant:
- Découpage du mot en : PRO + BASE1 + ENC.
- Découpage de *base1* en : PRE + BASE + SUF.
- Découpage de *base* en : racine + schème.

*2.4.1.3. Application de la technique*

Rappelons que le rôle de l'analyseur morphologique est d'extraire la racine des mots.

L'analyse morphologique de la langue arabe est plus complexe que celle de la plupart des langues européennes. On décompose l'analyse morphologique en trois phases principales:
- l'élimination des proclitique et des enclitiques
- l'élimination des préfixes et des suffixes
- l'extraction de la racine et du schème (وزن) correspondant si c'est possible ; sinon le mot est considéré comme un mot spécifique et par conséquent sa décomposition se limite aux deux phases précédentes.

Dans notre travail, on a essayé de construire un analyseur morphologique qui prend en charge la majorité des formes du lexique arabe (les verbes trilitères sains et les noms sains).

*2.4.1.3.1. Les étapes de découpage*

Comme mentionné précédemment, il existe trois étapes principales de découpage, il s'agit de:
- Découpage du mot en *proclitique +base1+enclitique* qui consiste à repérer tous les proclitiques et les enclitiques qui apparaissent dans le mot. Base1 est en général une base (racine + des infixes) munie de préfixes et de suffixes.
- Découpage de la base1 (résultat de la phase précédente) en *préfixes + base + suffixes*; le principe de cette phase est le même que celui de la phase précédente.
- Découpage de la base en *racine* et *schème* c'est-à-dire trouver un schème parmi les schèmes stockés dans le dictionnaire des schèmes qui corresponde à la base. La méthode de reconnaissance du schème sera décrite en II.4.2.

*2.4.1.3.2. Reconnaissance des proclitiques et des enclitiques*
**a) Notion de proclitique / enclitique**
   Il y a des particules qui s'additionnent au début ou à la fin d'un mot pour en changer le sens ou pour avoir un effet sur la rection du mot.

   Elles s'appellent les *enclises* (لواصق). Celles qui viennent au début de mot sont les *proclitiques* (لواصق قبلية) comme: «ل» (لام التوكيد), «ل» (لام الأمر), Comme dans les entités «ليضرب», «لكريم» ; et aussi le signe de la détermination «ال».

   Alors que celles qui viennent à la fin sont appelées *enclitiques* (لواصق بعدية), par exemple les pronoms affixes compléments comme (هم،ها) dans: (ضربهم،مجلسها).

**Remarque**
   Les proclitiques et les enclitiques ne sont pas libres d'apparaître au hasard, mais il existe une certaine compatibilité entre eux présentée table 3.

**b) Recherche des proclitiques et enclitiques**
   Fort heureusement la liste des proclitiques et des enclitiques de la langue arabe est limitée on peut utiliser la liste proposé par **[10]** plusieurs d'entre eux ont été utilisés par **[7]** pour la lemmatisation. Le découpage du mot en " proclitiques + base1+ enclitiques" ne se limite pas à la recherche d'un proclitique (respectivement un enclitique) parmi la liste au début (respectivement la fin) du mot, mais ainsi à une certaine compatibilité entre les proclitiques et les enclitiques repérés dans le mot à décomposer.

**Table 2:** liste des proclitiques et des enclitiques.

| Liste des proc | ' | ب | ك | ل | س ،ف | أ | ال | كال | لل | فب | فس | فال | فك | فل | فلل | أف | أس | فبال | فكال | بال |
|---|---|---|---|---|---|---|---|---|---|---|---|---|---|---|---|---|---|---|---|---|
| Liste des encl | ' | ه | ي | ك | هن | كم هما ها | كم كز | ني | ن | | | | | | | | | | | |

La table suivante présente les différents cas de compatibilité et d'incompatibilité entre les proclitiques et les enclitiques
- Les cases contenant le caractère correspondent à une incompatibilité du proclitique indiqué en tête de ligne et de l'enclitique indiqué en tête de colonne.
- Proclitique et enclitique compatibles donne une décomposition correcte.

**Table 3:** table de compatibilité entre proclitiques/ enclitiques**.**

| P\E | ه | ي | ك | هم | هن | هما | ها | كم | كن | كما | ذي | ذا |
|---|---|---|---|---|---|---|---|---|---|---|---|---|
| '' | | | | | | | | | | | | |
| ب | | | | | | | | | | | ○ | |
| ك | | | | | | | | | | | ○ | |
| ل | | | | | | | | | | | | |
| ف | | | | | | | | | | | | |
| س | | ○ | | | | | | | | | | |
| أ | | ○ | | | | | | | | | | |
| ال | ○ | ○ | ○ | ○ | ○ | ○ | ○ | ○ | ○ | ○ | ○ | ○ |
| بال | ○ | ○ | ○ | ○ | ○ | ○ | ○ | ○ | ○ | ○ | ○ | ○ |
| كال | ○ | ○ | ○ | ○ | ○ | ○ | ○ | ○ | ○ | ○ | ○ | ○ |
| لل | ○ | ○ | ○ | ○ | ○ | ○ | ○ | ○ | ○ | ○ | ○ | ○ |
| فب | | | | | | | | | | | ○ | |
| فس | | ○ | | | | | | | | | | |
| فال | ○ | ○ | ○ | ○ | ○ | ○ | ○ | ○ | ○ | ○ | ○ | ○ |
| فك | | | | | | | | | | | ○ | |
| فل | | | | | | | | | | | ○ | |
| فلل | ○ | ○ | ○ | ○ | ○ | ○ | ○ | ○ | ○ | ○ | ○ | ○ |
| أف | | | | | | | | | | | | |
| أس | | ○ | | | | | | | | | | |
| فبال | ○ | ○ | ○ | ○ | ○ | ○ | ○ | ○ | ○ | ○ | ○ | ○ |
| فكال | ○ | ○ | ○ | ○ | ○ | ○ | ○ | ○ | ○ | ○ | ○ | ○ |

**Test de compatibilité**

Après l'extraction du proclitique P et de l'enclitique E du mot à analysé, ces deux sous-chaînes sont fusionnés en une chaîne C pour les tester dans une table des incompatibilités (figure 7). Si la chaine C n'y est pas trouvée alors ce proclitique P est compatible avec cet enclitique E.

**Exemple**

L'analyse du mot (فبتعلمكم) donne comme proclitique (فب) et enclitique (كم), la fusion donne (فبكم); cette chaîne est compatible selon la table, alors le découpage est correct.

| Liste |
|---|
| بني |
| ………. |
| كالهما |
| ……… |
| فكالنا |

**Figure 7:** table des sous chaînes incompatibles.

**Remarque**
La table recense les incompatibilités, moins nombreuses que les compatibilités.
.
**c) Principe de l'analyse**

Lors du découpage du mot en proclitique+ base1 +enclitique le processus identifie le plus long proclitique (respectivement enclitique) du mot, puis il accède à la table pour vérifier la compatibilité entre les deux (proclitique et enclitique).

Si c'est compatible la décomposition est acceptée, elle sera stockée dans la table du résultat de cette phase, puis on continue avec une nouvelle décomposition pour traiter tous les cas possibles.

Sinon la décomposition est fausse, on passe à une autre décomposition.

**Exemple**

La décomposition du mot 'فكاتبهم' d'après le processus précédent donne

| Proclitique | ف |
|---|---|
| Enclitique | هم |
| base1 | كاتب |

**Figure 8 :** Décomposition en proclitiques enclitiques.

Qui est une décomposition correcte.

Malheureusement, ce n'est pas le cas pour tout le lexique arabe du fait qu'un verbe peut renfermer un radical (lettre de sa racine) comme un proclitique (respectivement un enclitique).

**Exemple**

La décomposition du mot 'فسمعهم' donne:

| proclitique | فس |
|---|---|
| enclitique | هم |
| base1 | مع |

**Figure 9 :** Décomposition erronée.

C'est une décomposition fausse malgré le mot 'مع' existe dans la langue arabe, car la racine de 'سمعهم' est 'سمع' et (non pas 'مع'). Le problème qui s'est posé dans ce cas est du au fait que 'س' est un radical de 'سمع' et un proclitique en même temps. La prise en compte de tous les cas envisageables de décomposition garantit la rencontre de la racine exacte du mot.

Malheureusement, ceci implique que le processus peut donner plusieurs racines, la solution est qu'une seule racine doit se retrouver dans le dictionnaire.

*2.4.1.3.3. Reconnaissance des préfixes et des suffixes*
**a) Notion de préfixe / suffixe**

Comme les enclises, les affixes sont concaténées au début (préfixes) ou à la fin (suffixes) du mot.

Les grammairiens arabes catégorisent les *préfixes* par les lettres qui sont ajoutées aux verbes pour exprimer l'*inaccompli* c'est-à-dire les lettres assemblées dans le mnémonique «انيت» (حروف المضارعة).

Les *suffixes* sont les lettres qui donnent une information sur le *genre* comme le «تاء» dans «الكريمة» (la généreuse) et le *nombre* «ان», «ات» dans «المجلسان, الكريمات » et les *pronoms affixes sujets* comme le «تم» dans «ضربتم».

Pour les préfixes et les suffixes, il existe aussi une certaine compatibilité qui va être illustrée dans une table.

**b) Principe de l'analyse**

Le principe de cette étape est pratiquement le même que la précédente sauf que la table de compatibilité utilisée est celle des préfixes et des suffixes ainsi que les lettres constituant les préfixes et les suffixes sont présenté dans la table suivante:

**Table 4:** liste des préfixes et des suffixes.

| Liste des préfixes | ا | ت | ن | ي | إ | ' ' | | | | | | | | | |
|---|---|---|---|---|---|---|---|---|---|---|---|---|---|---|---|
| Liste des suffixes | ' ' | ات | ية | ة | يات | نا | ت | تما | تم | تن | ن | ين | ان | ون | وا | ا | ي |

La table suivante est celle de la compatibilité entre les préfixes et les suffixes :

**Table 5:** table de compatibilité préfixes / suffixes.

| P\S | ' ' | ات | ية | يات | ت | تما | تم | تن | ن | ين | ان | ون | وا | ا | ي |
|---|---|---|---|---|---|---|---|---|---|---|---|---|---|---|---|
| ' ' | | | | | | | | | | | | | | | |
| ا | | ○ | ○ | ○ | ○ | ○ | ○ | ○ | ○ | ○ | ○ | ○ | ○ | ○ | ○ |
| ت | | ○ | ○ | ○ | ○ | ○ | ○ | ○ | | | | | | | |
| ي | | ○ | ○ | ○ | ○ | ○ | ○ | ○ | | ○ | | | | | ○ |
| ن | | ○ | ○ | ○ | ○ | ○ | ○ | ○ | ○ | ○ | ○ | ○ | ○ | ○ | ○ |
| إ | | ○ | ○ | ○ | ○ | ○ | ○ | ○ | | ○ | ○ | ○ | | | |

**Exemple**

La décomposition du mot 'فأعلن' d'après le processus de décomposition en préfixe+ base +suffixe et celui de proclitique+ base1 +enclitique donne :

| Proclitique | ف |
| Enclitique | ' ' |
| Base1 | أعلن |
| Préfixe | أ |
| Suffixe | ن |
| Base | عل |

**Figure 10:** Décomposition complète erroné.

La décomposition en *proclitique+ base1 +enclitique* est bonne, mais celle en *préfixe+ base +suffixe* ne l'est pas.

La solution proposée pour ce problème, comme dans la phase précédente, est de traiter tous les cas possibles; c'est-à-dire, pour un seul mot on doit engendrer tout un tableau de décompositions et de traiter les décompositions une par une jusqu'à ce qu'on arrive à une base correcte (racine existante).

**Exemple**

Découpage du mot 'فسأعلنه', la table suivante donne toutes les décompositions possibles, la base est à l'intérieur des cases et les cases vides représentent des cas interdits :

| | Pré(''), suffi('') | Pré('ا'), suffi('') | Pré(''), suffi('ن') | Pré('ا'), suffi('ن') |
|---|---|---|---|---|

| Proc(''), encl('') | | | | |
|---|---|---|---|---|
| Proc('فس') encl('') | أعلنه | علنه | | |
| Proc('ف') encl('') | سأعلنه | | | |
| Proc('') Encl('ه') | فسأعلن | | فسأعل | |
| Proc('ف') encl('ه') | سأعلن | | سأعل | |
| Proc('فس') encl('ه') | أعلن | **علن** | أعل | عل |

**Figure 11:** Exemple de découpage d'un mot.

De toutes les décompositions précédentes seule 'علن' est correcte puisque toutes les autres décompositions seront refusées dans la troisième partie (recherche de la racine et du schème), soit comparaison avec le dictionnaire des mots outils et spécifique, soit schème existant et racine introuvable (non entrée dans le dictionnaire).

Les deux bases (racine +schème) sont a priori correctes. Pour quelqu'un qui connaît la morphologie de l'arabe, la racine correcte est 'حسن' et non pas 'حس', mais pour un analyseur automatique les deux sont correctes.

Découpage du mot 'فسأحسنه'
-   Découpage1= [proc ('فس') ; encl ('ه') ; pré ('أ ') ; suffi (' ') ; base ('حسن')].
-   Découpage2= [proc ('فس') ; encl ('ه') ; pré ('أ') ; suffi ('ن ') ; base ('حس')].

Les deux bases (racine +schème) sont correctes mais pour quelqu'un qui connaît la morphologie de l'arabe, la racine correcte est 'حسن' et non pas 'حس', mais pour un analyseur automatique les deux sont correctes.

*2.4.2. Recherche de schème et de la racine*
*2.4.2.1. Méthode de recherche de schème*
Le principe de la méthode de recherche est très simple. Pour un mot X, un schème i du dictionnaire des schèmes correspond au mot X si la taille de schème est égale a la taille de mot X et si toutes les lettres correspondantes aux positions dans le champ listeinfixe se trouvent dans le mot X aux mêmes positions révélées par ce champ (listeinfixe).

Voici un exemple qui nous aide à comprendre le processus:

Mot = 'صالح' le processus de recherche de schème parcourt tous les enregistrements qui ont la même taille avec le mot jusqu'à la rencontre du schème

'فاعل'. Le champ listeinfixe correspondant est '2 ' la lettre 'ا' trouve à la position 2 du mot 'صالح' donc c'est probablement le bon schème.

*2.4.2.2. Recherche de la racine*

Après la détermination du schème, l'extraction de la racine se limite à la suppression de toutes les lettres correspondantes aux positions de champs listeinfixe dans le mot à décomposer.

**Exemple**

Le mot 'صالح' a pour schème 'فاعل', le champ listeinfixe est '2'.
L'élimination de la lettre 'ا' de la position 2 du mot 'صالح' qui est la même du champ listeinfixe donne 'صلح'. Ainsi on a retrouvé la racine correcte du mot 'صالح' qui est la racine 'صلح'

*2.4.3. Interprétation*

Après extraction de la racine et recherche lexicale, on procède à la génération des valeurs morphologiques du mot, ces informations seront utiles dans le module d'analyse syntaxique qui sert a extraire les syntagmes pertinents.

**Exemple**

(مكتوب) à pour schème (مفعول) dans le champ catégorie du schème (مفعول) on trouve la catégorie (nom) on affecte cette catégorie au mot (مكتوب).

*2.4.4. Interprétation*

Ces trois étapes ne sont pas les seules mais il y a d'autres étapes qui peuvent être introduites entre elles en cas de besoin (traitement de compatibilité entre les proclitiques et enclitiques et préfixes et suffixes, désambiguïsation dans le cas où il y a plusieurs interprétations, extraction de la racine).

Ces valeurs morphologiques sont extraites du dictionnaire des schèmes qui fait correspondre à chaque schème une catégorie grammaticale

**Exemple**

    مفاعلة: nom
    انفعل: verbe
    فاعل: verbe,nom.

## 3. Conclusion

L'utilisation de la langue arabe comme moyen de communication à travers le support informatique a été longtemps appréhendé avec beaucoup d'hésitation par la communauté scientifique, particulièrement celle du monde arabe où cet outil trouve beaucoup d'utilisations importantes.

Nous avons entrepris ce travail pour montrer que cette tentative est faisable, et aussi pour relever le défi visant à offrir des outils informatiques pour l'analyse des documents en langue arabe. En effet nous avons montré que les études théoriques issues du Traitement Automatique des Langues Naturelles s'adaptent bien au traitement de la langue arabe avec quelques spécificités propres.

L'étude morpho lexicale nous a permis de pouvoir : segmenter le mot en préfixes, postfixes et base et extraire le schème et la racine (qui veut dire reconnaissance de toutes les formes fléchies que peut prendre un mot, sans la connaissance à priori de ces mots) et donner toutes les valeurs morphologiques associées au mot.

Bien que la langue arabe soit aussi riche que vaste, nous pensons que notre présente étude apporte quelques solutions qui sont exploitables dans l'immédiat.

**Références**
**[1]** Ali, Nabil: 2003, *The Second Wave of Arabic Natural Language Processing.* (2003).
**[2]** Aljlayl, Mohamed. And Ophir, Frieder: 2002, *On Arabic Search: Improving the Retrieval Effectiveness via a Light Stemming Approach,* In 11th International Conference on Information and Knowledge Management (CIKM), November (2002).
**[5]** Attia, Mohamed, *A.*: 2004, *Report on the introduction of Arabic to ParGram.* The ParGram Fall Meeting, National Centre for language Technology, Ireland, (2004).
**[3]** Attia, Mohamed, A.: 2000 *A large-scale computational processor of the Arabic morphology*, A Master's Thesis, Cairo University, (Egypt) (2000).
**[4]** Ttia, Mohamed, A.: 2005, *Devloping Robust Arabic Morphological Transducer Using Finite State Technology* In 8 th annual CLUK Research Colloquium (2005).
**[6]** Chalabi, Achraf: 2000, *MT-Based Transparent Arabization of the Internet TARJIM.COM. In White*, J.S.(Ed) AMTA (2000) Springer: Verlag Berlin Heidelberg,(2000)
**[7]** Chen, A. And. Gey, F.: 2002, *Building an Arabic Stemmer for Information Retrieval.* Proceedings of the Eleventh Text REtrieval Conference (TREC 2002). National Institute of Standards and Technology, Nov 18-22, (2002).
**[8]** Daimi, Kevin: 2001, *Identifying Syntactic Ambiguities in Single-Parse Arabic Sentence. In Computer and humanities* (2001).
**[9]** Darwish, K. And Oard, D.: 2002, *CLIR Experiments at Maryland for TREC 2002: Evidence Combination for Arabic-English Retrieval in TREC*, Gaithersburg, MD (2002).
**[10]** Darwish, K.: 2003, *Probabilistic Methods for Searching OCR-Degraded Arabic Text*, Doctoral dissertation, University of Maryland, (2003).
**[11]** Douzidia, Fouad, Soufiane, Guy: 2005, *Lapalme Un système de résumé de textes en arabe* 2ème Congrès International sur l'Ingénierie de l'Arabe et l'Ingénierie de la langue, Alger, (2005).
**[12]** Groblink, M., Mladenic, D.: 2004, *Text mining tutorial* Slovenia (2004)

**[13]** Hassoun, M.O.: 1989, *Conception d'un dictionnaire pour le traitement automatique de l'arabe dans différents contextes d'applications,* Thèse de doctorat, Université Lyon 2, (1989)

**[14]** Larkey, L.S., Ballesteros, L. And Connell, M.: 2002, *Improving Stemming for Arabic Information Retrieval: Light Stemming and Co-occurrence Analysis*, In Proceedings of the 25th Annual International Conference on Research and Development in Information Retrieval (SIGIR 2002), Tampere, Finland, August (2002).

**[15]** Lasakri, Mohamed, Taib: 1994, *Sémantique du langage naturel à travers un système support de thésaurus* Thèse d'état, (1994)

**[16]** Lehmam, A. et Bouvet, P.: 2004, *Un résumé de textes, application à la langue arabe.* TALN (2004)

**[17]** Mohamadi, T.S. Mokhnache: 2002, *Design and development of Arabic speech synthesis*, WSEAS 2002, Greece, Sept. 25-28, (2002).

**[18]** Selmane, S. & Zergoug, D.: 1995, *Correcteur d'erreurs de frappe d'un éditeur de textes arabes non voyellés Alger (*1995)

**[19]** Taibi, Nacera: 1997, *Contribution a l'étude du traitement automatique des erreurs dans un texte écrit en arabe* Thèse de magister ENS (SH) Alger 1997

**[20]** Zemirli, Zouhir: 1996, *Un analyseur destiné à l'aide à la construction d'une base de donnes lexicales de la langue arabe.*Colloque international "Langues situées, technologie et communication" IERA, (1996)